\documentclass{bmvc2k}

\usepackage{amssymb}
\usepackage{bm}
\usepackage{float}
\usepackage{multirow}
\usepackage{comment}
\usepackage[dvipsnames,table]{xcolor}
\usepackage{tabularx}
\usepackage{xspace}
\usepackage{booktabs}
\usepackage{subfigure}
\usepackage{wrapfig}

\definecolor{first}{RGB}{255,178,178}
\definecolor{second}{RGB}{255,217,178}
\definecolor{third}{RGB}{255,255,178} 

\def\ours{Ev4DGS}

\newcommand{\argmin}{\mathop{\rm arg~min}\limits}
\newcommand{\argmax}{\mathop{\rm arg~max}\limits}

\newcolumntype{C}{>{\centering\arraybackslash}X}

\makeatletter
\DeclareRobustCommand\onedot{\futurelet\@let@token\@onedot}
\def\@onedot{\ifx\@let@token.\else.\null\fi\xspace}

\def\eg{\emph{e.g}\onedot} 
\def\ie{\emph{i.e}\onedot}

\makeatother

\usepackage[capitalize]{cleveref}
\crefname{section}{Sec.}{Secs.}
\Crefname{section}{Section}{Sections}
\Crefname{table}{Table}{Tables}
\crefname{table}{Tab.}{Tabs.}

\title{Ev4DGS: Novel-view Rendering of Non-Rigid Objects from Monocular Event Streams}

\addauthor{Takuya Nakabayashi}{nakka0204@keio.jp}{1}
\addauthor{Navami Kairanda}{nkairand@mpi-inf.mpg.de}{2}
\addauthor{Hideo Saito}{hs@keio.jp}{1}
\addauthor{Vladislav Golyanik}{golyanik@mpi-inf.mpg.de}{2}

\addinstitution{
 Keio University\\
 Yokohama, Japan
}
\addinstitution{
 Max Planck Institute for Informatics\\
 Saarbrücken, Germany
}

\runninghead{Nakabayashi et al.}{Ev4DGS}

\def\eg{\emph{e.g}\bmvaOneDot}

\begin{document}

\maketitle

\begin{abstract}

Event cameras offer various advantages for novel view rendering compared to synchronously operating RGB cameras, and efficient event-based techniques supporting rigid scenes have been recently demonstrated in the literature. 
In the case of non-rigid objects, however, existing approaches additionally require sparse RGB inputs, which can be a substantial practical limitation; it remains unknown if similar models could be learned from event streams only. 
This paper sheds light on this challenging open question and introduces Ev4DGS, i.e.,~the first approach for novel view rendering of non-rigidly deforming objects in the explicit observation space (i.e., as RGB or greyscale images) from monocular event streams. 
Our method regresses a deformable 3D Gaussian Splatting representation through 1) a loss relating the outputs of the estimated model with the 2D event observation space, and 2) 
a coarse 3D deformation model trained from binary masks generated from events. 
We perform experimental comparisons on existing synthetic and newly recorded real datasets with non-rigid objects. 
The results demonstrate the validity of Ev4DGS and its superior performance compared to multiple na\"{i}ve baselines that can be applied in our setting. 
We will release our models and the datasets used in the evaluation for research purposes; see the project webpage: \url{https://4dqv.mpi-inf.mpg.de/Ev4DGS/}. 
\end{abstract}
\section{Introduction}
\label{sec:intro}

Non-rigid 3D reconstruction of deforming objects has a wide range of applications \textit{e.g.,} human avatar generation, virtual/augmented reality, and automatic robot manipulation.
These applications often require multiple cameras, making long-term camera calibration and large equipment indispensable.
If similar tasks could be performed with a single camera, it would lead to the human avatar generation with a single smartphone and downsizing of the robot.
However, this problem is severely ill-posed because the non-rigid object's shape changes over time, though an observation from only one viewpoint is available per one timestamp.
Despite significant advances, the problem remains unsolved, and there is a long way to go until monocular techniques can offer a reasonable alternative to other sensor types in terms of reconstruction quality. 

Neural Radiance Fields (NeRF) \cite{mildenhall2021nerf} and 3D Gaussian Splatting (3DGS) \cite{kerbl3Dgaussians} are breakthroughs in the field of 3D reconstruction and have fueled the field of novel-view rendering for dynamic scenes.
Recent works \cite{pumarola2021d, tretschk2021non, park2021nerfies, park2021hypernerf, cao2023hexplane, yang2024deformable, liang2023gaufre, kratimenos2024dynmf, lu20243d, lin2024gaussian, das2024neural, yu2024cogs, huang2024sc, wu20244d, yang2023gs4d} extend them with deformation fields or 4D extension
to address this problem and 
support the monocular non-rigid 3D reconstruction and novel-view synthesis. 
Although previous methods take RGB images as input, the RGB-based monocular setting has disadvantages.
Motion blur that occurs when the camera or object moves at high speed can degrade rendering performance \cite{ma2022deblur}. 
In addition, AR glasses and drones are limited in the amount of power they can use, and cannot be equipped with RGB cameras that consume high amounts of power.

Event cameras are promising and reasonable alternatives to RGB cameras in these scenarios. 
An event camera is a sensor inspired by the visual system of animals that records asynchronous brightness changes in a scene---called events---instead of RGB frames at regular intervals. 
Event cameras are characterized by high temporal resolution, enabling measurement of scenes involving fast motion that are difficult to capture with RGB cameras.
While there are works addressing the novel-view synthesis of static scenes with event streams  \cite{rudnev2023eventnerf, hwang2023ev, klenk2023nerf, wu2024ev, xiong2024event3dgs, low2023robust}, 
so far, no solution to non-rigid 3D reconstruction for a single monocular event camera has been demonstrated. 

We demonstrate that a solution to this challenging problem is possible. 
Inspired by the recent advances in RGB-based reconstruction \cite{novotny2019c3dpo, johnson2023ub4d, ma2023deformable, das2024neural}, this paper proposes Event-based 4D Gaussian Splats \textbf{(Ev4DGS)}, the first method for the novel-view rendering of non-rigid objects from monocular event streams recorded by a continuously moving event camera. 
Our main observation is that event-based supervision can be combined with the estimation of 3D shapes serving as a basis for a regularized low-rank deformation model.
However, there are substantial differences between the RGB- \cite{das2024neural} and event-based settings (ours) following similar principles.
We define loss functions that can be computed using events and binary masks that can be generated from events, allowing us to train the whole framework using event streams only. 
In the first stage, the motion is tracked with a coarse deformation model and in the second stage, the object's appearance is represented with 4D Gaussians \cite{wu20244d, yang2023gs4d}, i.e.,~generalisation of the state-of-the-art 3DGS technique for the novel-view rendering of rigid scenes to the temporal domain.
The two-stage training process allows us to stabilize the training by reducing the number of optimised parameters in each stage (and cope with smaller ill-posed problems).
Our deformation model is a time-dependent linear combination of low-rank basis point clouds: we relate the obtained basis shapes with the temporal changes of the 3D Gaussian parameters. 
The closest work \cite{ma2023deformable} to ours that tackles novel-view synthesis of dynamic scenes also requires RGB image input, which can be a practical limitation. 
Since no datasets exist that fulfil the requirements of the new evaluation setting, we render and record synthetic and real datasets allowing quantitative and qualitative evaluation.
In summary, the primary \textbf{technical contributions} of this paper are as follows: 
\begin{itemize} 
    \item Ev4DGS, the first approach for novel-view rendering of non-rigid scenes from monocular event cameras. 
    The proposed framework enables self-supervised learning with only event streams and camera tracking information; no additional inputs are required. 
    \item We employ a coarse deformation model to represent large scene displacements tailored to the event-based setting. 
    This model can be trained from monocular observations despite the ill-posedness of the problem. 
    \item New synthetic and real sequences for evaluating Ev4DGS and future methods operating under similar assumptions. 
\end{itemize} 

In our experiments, Ev4DGS demonstrates state-of-the-art accuracy, higher on average than that of competitive baselines. 
At the same time, it does not require an explicit reconstruction of RGB images. 
\section{Related Work}
\label{sec:related_work}
At present, RGB-based methods are the most common method for novel-view synthesis of dynamic scenes. 
In this section, we introduce several RGB-based methods and then introduce related event-based methods.

\subsection{Monocular RGB-based Dynamic Novel-view Synthesis}
\label{ssec:related_work_rgb_based_methods}
The development of deep learning approaches in the last decade has brought significant progress in novel-view synthesis.
Neural Radiance Fields (NeRF) \cite{mildenhall2021nerf} is one of the most significant breakthroughs.
NeRF is a field learned by MLP that stores the colour and density that depends on the viewing direction at each 3D location.
Approaches to adapt NeRF, which performs 3D reconstruction of static scenes, to dynamic scenes have been actively studied \cite{pumarola2021d, tretschk2021non, park2021nerfies, park2021hypernerf, cao2023hexplane}. 
One of the primary drawbacks of NeRF-based methods is the long training and rendering time.

3D Gaussian Splatting (3DGS) \cite{kerbl3Dgaussians} achieves real-time rendering while maintaining high rendering quality.
Unlike NeRF, 3DGS, with its explicit 3D representation, allows for more direct motion representation.
The approaches for representing motions in existing 3DGS methods for dynamic scenes include the combination of the canonical space and the deformation field \cite{yang2024deformable, liang2023gaufre, kratimenos2024dynmf, lu20243d, lin2024gaussian, das2024neural, yu2024cogs, huang2024sc}, and 4D Gaussian \cite{wu20244d, yang2023gs4d}.
In this paper, we combine a canonical space and a deformation field to represent motion.
The canonical space is a set of static 3D Gaussians that represent the 3D structure in the scene, and the deformation field.
In particular, we employ a deformation inspired by the coarse deformation model proposed in NPGs \cite{das2024neural}.
In NPGs, non-rigid object shapes are represented as time-varying coarse point clouds, and each 3D Gaussian is anchored to a local volume formed by the coarse points.
Since our method differs from NPGs in that it takes events as input, we add several mechanisms to accommodate this.

\subsection{Event-based Novel-view Synthesis}
\label{ssec:related_work_event_based_methods_for_rigid_scenes}

Most event-based methods target rigid scenes. 
Only recently, methods for non-rigid scenes emerged.

\paragraph{Rigid Scenes.} 

The event camera is not affected by motion blur due to its high temporal resolution.
Existing methods \cite{qi2023e2nerf, qi20243, yu2024evagaussians, deguchi2024e2gs} perform novel-view synthesis using both event streams and RGB images with motion blur.
They formulate the relationship between the motion blur in RGB frames and events and incorporated it into the loss function.

Recently, methods~\cite{rudnev2023eventnerf, hwang2023ev, klenk2023nerf, wu2024ev, xiong2024event3dgs, low2023robust} that are less restrictive with regards to input and directly operated on event streams are proposed.
Since they cannot directly supervise the rendered frame, they introduce a loss function that compares the brightness change estimated from two rendered images with that estimated from an event stream, and it enables learning of NeRF and 3D Gaussians with only event streams as input.

\paragraph{Non-rigid Scenes.} 
A few works tackle novel-view synthesis of dynamic scenes using event cameras.
EvDNeRF \cite{bhattacharya2024evdnerf} is an event camera simulator that uses known event streams to train a network with D-NeRF \cite{pumarola2021d} as the backbone, allowing event streams at novel viewpoints to be output.
DE-NeRF \cite{ma2023deformable} works on novel-view synthesis of dynamic scenes using event streams and RGB images as input, however, it also requires RGB image input.
To the best of our knowledge, there is still no research that addresses the novel-view synthesis of dynamic scenes using only monocular event streams as input.

\section{Preliminaries}
\label{sec:preliminaries}

We next review the event formation model as we seek to reconstruct from dynamic event stream. 
Event cameras record the asynchronous per-pixel brightness changes---observed from their viewpoint---as events.
An event $e$ is thus represented as a tuple $(x,y,t,p)$, where the 2D coordinates $(x,y)$ indicate the event position in the image plane, and $t$ denotes the timestamp when the event occurs, and the polarity $p$ is a binary value representing whether the pixel becomes brighter or darker.
An event is triggered at $t$ when a logarithmic intensity change from $t'$ exceeds a certain threshold $\sigma$, where $t'$ is the timestamp corresponding to the previous event occurring at $(x,y)$.
If we denote this log intensity change as $L_{x,y}(t)$, the polarity of the captured event is determined as follows:
\begin{equation}
    p  = \left\{
        \begin{array}{ll}
        +1, & \text{if }  L_{x,y}(t)-L_{x,y}(t')>\sigma\\
        -1, & \text{if }  L_{x,y}(t)-L_{x,y}(t')<-\sigma.
        \end{array}
        \right.
\end{equation}
Therefore, we can formulate the intensity change (aka \emph{accumulated difference}) between two timestamps $t_0$ and $t$ as follows:
\begin{equation}
    L_{x,y}(t)-L_{x,y}(t_0)=\sigma\sum_{t_0 < t_i \leq t} p_i \triangleq E_{x,y}(t_0,t).
    \label{eq:event_accumulation}
\end{equation}

\section{Method}
\label{sec:method}
\begin{figure*}
    \centering
    \includegraphics[width=\linewidth]{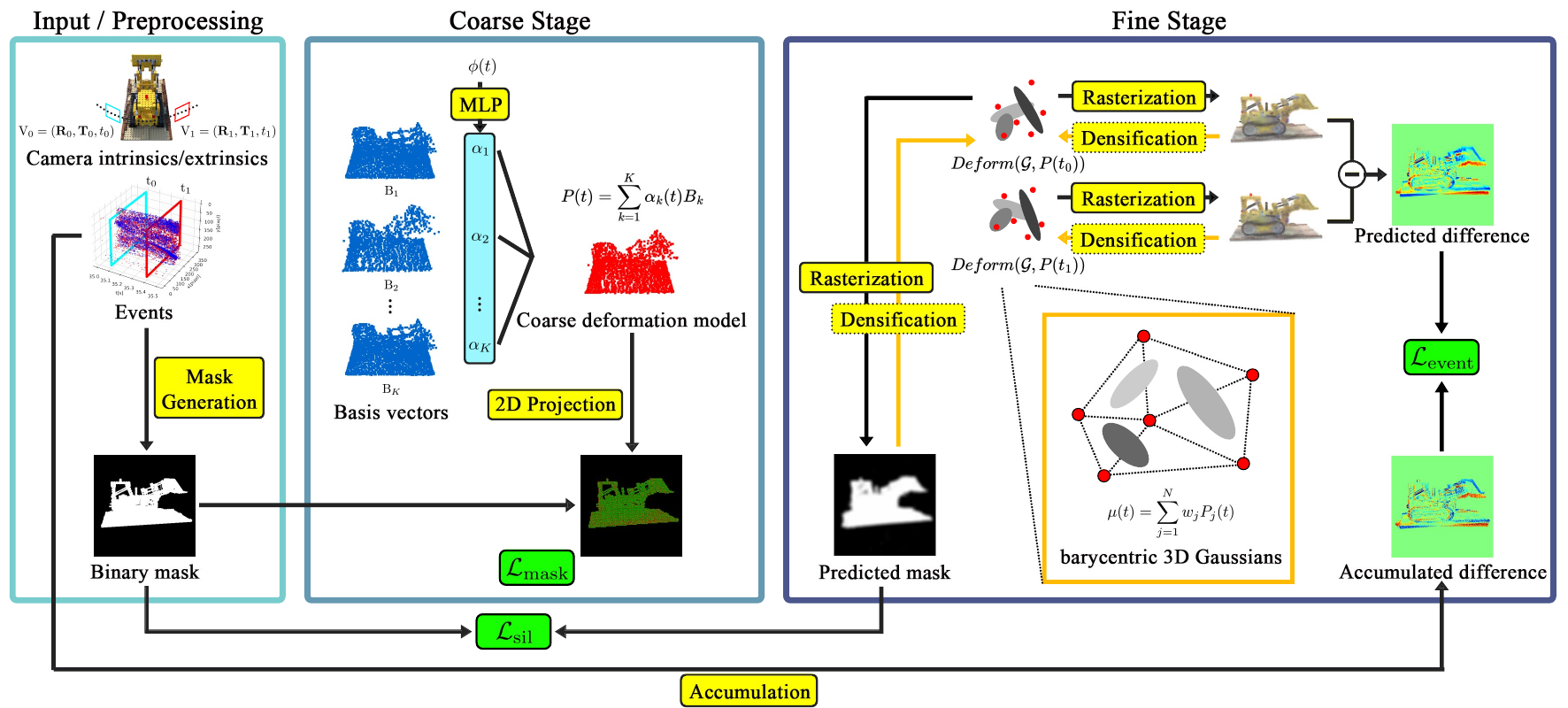}
\caption{Overview of the proposed Ev4DGS method. We divide the event-based reconstruction of non-rigid scenes into two stages. In the first stage, we train the coarse deformation model, which represents a non-rigid object shape as a set of points and enables 
the representation of large 3D deformations in a scene. In the second stage, we obtain the 3DGS representation from an event stream.
Best viewed with zoom. 
}
    \label{fig:overview}
\end{figure*}
We propose~\ours, a new method for novel-view synthesis of non-rigidly deforming objects from a monocular event stream.
We assume that the event camera moves around the non-rigid object rapidly, and in addition to the event stream, the viewpoints $\{\mathbf{V}_i\}_{i=1}^{N_v}$ corresponding to the timesteps $\{t_i\}_{i=1}^{N_v}$ are known in the training.
\emph{Although we only observe each deformed state from a single viewpoint and with sparse event data at training, we can query any view and at any time as intensity images during inference.}
Moreover, we make no assumptions about the shape (\eg, template) or motion (\eg, rigidity) of the object.

Accurate 3D reconstruction of deformable objects from monocular observations is severely ill-posed. 
An end-to-end learning of high-dimensional representations such as a deformable 3DGS can be prone to the local optima.
To simplify this process, we factorise the non-rigid object's representation as well as the self-supervised learning into its motion and appearance.
More specifically, we train independently in two stages and integrate them to represent dynamic scenes.
In the first stage, we train a coarse deformation model to represent large deformations in the scene~(\cref{sssec:stage1}). 
In the second stage, we obtain the full 3DGS representation of the non-rigid object for the dynamic novel-view synthesis~(\cref{sssec:stage2}).
An overview of our proposed method is shown in \cref{fig:overview}.

\noindent \textbf{Mask Generation.}
The inputs to \ours~are a monocular event stream and calibrated camera parameters. 
As part of our method, we also require binary segmentation masks of the object to compute the mask and silhouette losses. 
We generate these segmentations $\{\hat{\mathbf{M}}\}_{i=1}^{N_v}$ directly from the event stream by simply applying Snakes \cite{kass1988snakes} to an image mapping the location of the event.

\subsection{Coarse Stage: Deformation Model} 
\label{sssec:stage1}
\noindent
\textbf{Parametric Point Representation.}
Inspired by NPGs \cite{das2024neural}, we employ a coarse model to represent the object deformations.
The model represents the evolving state of the object as a set of time-dependent coarse points, $\mathbf{P}(t) \in \mathbb{R}^{N_c \times 3}$, where $N_c$ is small ($\approx 1000$).
The point model $\mathbf{P}(t)$ is computed as the linear combination of learned low-rank deformation basis vectors $\{\mathbf{B}_k \in  \mathbb{R}^{N_c \times 3} \}_{k=1}^K$ as follows:
\begin{equation}
    \mathbf{P}(t)=\sum_{k=1}^K \alpha_k(t) \mathbf{B}_k,
\end{equation}
where $K$ is the number of basis vectors and $\alpha_k(t)$ are the time-dependent basis weights. 
Following earlier neural parametric models~\cite{palafox2021npms,das2024neural}, the weights are learned from an MLP $f_\theta:\phi(t)\mapsto\{\alpha_k\}_k$ with sinusoidal positional encoded time $\phi(t)$ as input.
Here, $K$ affects the rigidity of the model, \ie, when $K$ is small, the model has fewer degrees of freedom, allowing for a more rigid modelling.
Next, we describe the losses that drive the optimisation of the basis $\{\mathbf{B}_k\}_{k=1}^K$ and MLP weights $\theta$. 

\noindent
\textbf{Mask Loss.}
An accurately reconstructed deformed point model will perfectly project to the interior of the input masks.
Given a 2D point set $\mathbf{P}_{\hat{\mathbf{M}}_i}$ uniformly sampled from the binary mask $\hat{\mathbf{M}}_i$, and the projection $\pi(.)$ of $\mathbf{P}(t_i)$ onto the image plane, we define the mask loss as follows: 
\begin{equation}
    \mathcal{L}_\mathrm{mask}=\sum_{i=1}^{N_v} CD(\mathbf{P}_{\hat{\mathbf{M}}_i},\pi(\mathbf{P}(t_i))),
\end{equation}
where $CD(\cdot,\cdot)$ is the Chamfer distance between two point clouds.
By minimising $\mathcal{L}_\mathrm{mask}$, the model learns the instantaneous shape of the non-rigid object.

\noindent
\textbf{Optimisation.}
The optimal MLP weights $\hat{\theta}$ and the basis vectors $\{\hat{\mathbf{B}}_k\}_{k=1}^K$ are obtained as:
\begin{equation}
    \hat{\theta},\{\hat{\mathbf{B}}_k\}_{k=1}^K=\argmin_{\theta,\{\mathbf{B}_k\}_{k=1}^K} \mathcal{L}_\mathrm{mask}.
\end{equation}

\subsection{Fine Stage: Dynamic Novel View Synthesis}
\label{sssec:stage2}
\noindent
\textbf{3D Gaussian Model.}
Inspired by NPG~\cite{das2024neural}, we employ a local volume driven Gaussian model $\mathcal{G}(t_i)$, where a Gaussian centre is computed with barycentric interpolation as:
\begin{equation}
    \boldsymbol{\mu}(t_i) = \sum_{j=1}^k w_{j} \mathbf{P}_j(t_i)
\end{equation}
where $k$ is the number of neighbours in the local volume, $ \mathbf{w} = \{w_j\}_{j=1}^{k}$ are barycentric weights and $\mathbf{P}_j(t_i)$ is $j$-th neighbour coarse point.
We optimise the full set of Gaussian parameters, \ie 
$\mathcal{G}(t_i) = \{ \mathbf{w}_i, \mathbf{R}_i, \mathbf{S}_i, \mathbf{o}_i, c_i \}_{i=1}^{N_g}$.
Note that Gaussian parameters are shared over time and driven fully by the coarse deformation model. 

\noindent
\textbf{Event Loss.}
We employ the event loss~\cite{rudnev2023eventnerf} to optimize 3D Gaussian parameters from the event streams.
First, we render the Gaussians~\cite{kerbl3Dgaussians} from two spatially and temporally close viewpoints $\mathbf{V}_i$ and $\mathbf{V}_j$, where $L(t_i) = L(\mathcal{R}(\mathcal{G}(t_i),\mathbf{V}_i))$ converts the rasterised pixel intensity to the logarithmic scale with gamma correction, similar to previous works~\cite{rudnev2023eventnerf}.
Further, we compute the accumulated difference $E(t_i,t_j)$ from the input event stream~(\cref{eq:event_accumulation}) and compare it to the predicted brightness change.
Then, the method learns the unknown model parameters in a self-supervised manner by minimising the $\mathcal{L}_\mathrm{event}$, defined as follows:
\begin{equation}
\begin{split}
    \mathcal{L}_\mathrm{event}&=\frac{1}{N_v}\sum_{i=1}^{N_v} \| (L(t_j)-L(t_i)) - E(t_j,t_i)\|_1,\\
    t_j &\sim U[t_i, t_i + L_\mathrm{max})
    \label{eq:event_loss}
\end{split}
\end{equation}
where we evaluate the loss in randomly sampled time windows~\cite{rudnev2023eventnerf}, with $L_\mathrm{max}=0.02$. 
Importantly, the event loss is computed over all pixels, \ie for each pixel where an event occurs and where it doesn't occur.
Additionally, we apply the colour filter mask to this loss function when the input is a colour event stream, similar to previous works~\cite{rudnev2023eventnerf}.

\noindent
\textbf{Silhouette Loss.}
In contrast to RGB images, events are spatially sparse and require additional supervision compared to RGB-based methods~\cite{das2024neural}.  
To suppress unnecessary Gaussians present in the background, we add silhouette loss $\mathcal{L}_\mathrm{sil}$ inspired by $\phi$-SfT~\cite{kair2022sft}.
We render the predicted binary mask $\mathbf{M}_i$ corresponding to the viewpoint $\mathbf{V}_i$ with the Gaussian rasterizer.
Thus, $\mathcal{L}_\mathrm{sil}$ can be computed as follows:
\begin{equation}
    \mathcal{L}_\mathrm{sil}=\sum_{i=1}^{N_v} \| \mathbf{M}_i-G(\hat{\mathbf{M}}_i) \|_{1},
\end{equation}
where $G(\cdot)$ is a function to apply Gaussian filter to the input mask and $\hat{\mathbf{M}}_i$ is a binary mask generated from event streams with Snakes. 

\noindent
\textbf{Optimisation.}
We formulate the optimisation of a deformable 3DGS representation as follows:
\begin{equation}
    \hat{\mathcal{G}},\hat{\mathbf{\theta}},\{\hat{\mathbf{B}}_k\}_{k=1}^K=\argmin_{\mathcal{G},\theta,\{\mathbf{B}_k\}_k} (\lambda_e \mathcal{L}_\mathrm{event}+\lambda_s \mathcal{L}_\mathrm{sil}).
\end{equation}
where $\lambda$s are the respective loss weights.
While the above optimisation primarily learns the Gaussian parameters, it additionally fine-tunes the MLP weights and the basis vectors from the coarse stage. 
\section{Experiments}
\label{sec:experiments}

\subsection{Datasets}
\label{ssec:dataset}
Since existing dataset \cite{ma2023deformable} of dynamic monocular event streams
lack variation in the shape and motion of the object,
we prepare new synthetic and real datasets to evaluate our proposed method quantitatively and qualitatively.

\noindent
\textbf{Synthetic Dataset.}
The dataset includes three sequences of different object shapes performing varied motions: Dynamic Lego \cite{ma2023deformable}, Jumping Jacks \cite{jumping_jacks_dataset}, and T-Rex \cite{t_rex_dataset}.
To simulate events, we use ESIM \cite{rebecq2018esim}.
ESIM accepts an RGB video input, and it outputs an event stream of the scene as if it was captured by an event camera.
Therefore, we first render an animation in 4k fps,
in which the camera moves continuously around deformable objects, with Blender.
The rendered animation is then provided as input to ESIM for event generation.
We set the threshold $\sigma$ for event triggering to $0.2$.

\noindent
\textbf{Real Dataset.}
We also create the dataset captured with a real event camera to evaluate our method on real data.
The recording process is the same as EventNeRF \cite{rudnev2023eventnerf}.
We place and fix an excavator toy on a turntable, and record it rotating with DAVIS MONO 346 and GoPro RGB camera.
Event streams recorded by DAVIS are used for training, and GoPro camera captures non-blurry GT frames for test. 

\begin{figure*}[h]
    \centering
    \includegraphics[width=\linewidth]{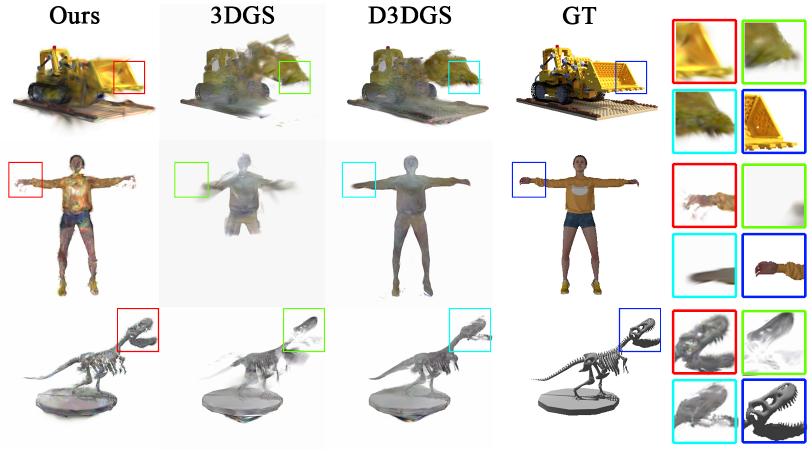}
    \caption{Qualitative results on the synthetic dataset (novel views). Our Ev4DGS outputs high-quality novel-view rendering as expected in this challenging setting. Competing methods miss object parts and details in the object interiors.}
    \label{fig:qualitative_result_synthetic}
\end{figure*}
\begin{figure*}
    \centering
    \includegraphics[width=\linewidth]{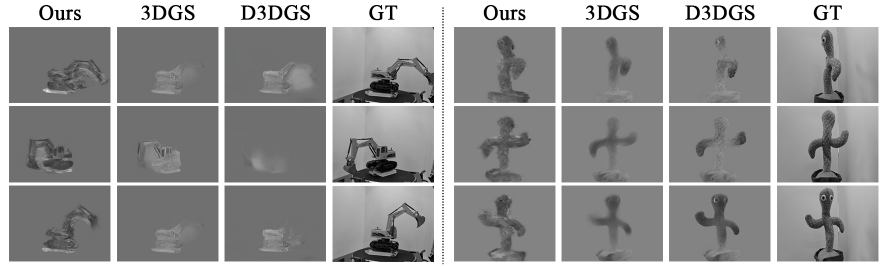}
    \caption{Qualitative results on our real dataset. Our Ev4DGS synthesises the spatially (shape, texture) and temporally coherent novel views that match the GT test views, while the competing methods fail.} 
    \label{fig:qualitative_result_real}
\end{figure*}
\begin{table*}[h!]    
    \small
    \centering
    \begin{tabularx}{\textwidth}{ccCCC}
        \toprule
        Sequence & & 3DGS & D3DGS & Ours\\ \midrule
        \multirow{2}{*}{Dynamic Lego} & PSNR$\uparrow$ & \cellcolor{third}16.692 & \cellcolor{second}17.645 & \cellcolor{first}21.410\\
         & SSIM$\uparrow$ & \cellcolor{third}0.807 & \cellcolor{first}0.845 & \cellcolor{first}0.845\\ \midrule
        \multirow{2}{*}{Jumping Jacks} & PSNR$\uparrow$ & \cellcolor{third}16.506 & \cellcolor{second}18.812 & \cellcolor{first}19.712\\
         & SSIM$\uparrow$ & \cellcolor{second}0.877 & \cellcolor{first}0.891 & \cellcolor{third}0.875\\ \midrule
         \multirow{2}{*}{T-Rex} & PSNR$\uparrow$ & \cellcolor{third}18.725 & \cellcolor{second}20.428 & \cellcolor{first}20.846\\
         & SSIM$\uparrow$ & \cellcolor{third}0.886 & \cellcolor{first} 0.898 & \cellcolor{second}0.888\\ \midrule \midrule
          \multirow{2}{*}{Average} & PSNR$\uparrow$ & \cellcolor{third}17.326 & \cellcolor{second}18.962 & \cellcolor{first}20.656\\
         & SSIM$\uparrow$ & \cellcolor{third}0.857 & \cellcolor{first}0.878 & \cellcolor{second}0.869\\
         \bottomrule
    \end{tabularx}
    \caption{Quantitative evaluation results on the synthetic dataset. 
    While Deformable 3DGS~\cite{yang2024deformable} marginally outperforms our approach on average on SSIM, ours outperforms Deformable 3DGS by $9\%$ on PSNR while not requiring explicit grayscale images. Red: best score, Orange: second best score, Yellow: third best score.
    }
    \label{tab:quantitative_results_on_synthetic_dataset_full}
\end{table*}
\begin{table}[t]
    \begin{minipage}[t]{.5\textwidth}
    \centering
    \begin{tabular}{l|cc}
        \toprule
         & PSNR$\uparrow$ & SSIM$\uparrow$\\ \midrule
        Ours w/o $\mathcal{L}_\mathrm{sil}$ & 20.222 & 0.859\\
        Ours w/o finetuning & 19.797 & 0.853\\
        Ours (Full) & \textbf{20.656} & \textbf{0.869}\\
         \bottomrule
    \end{tabular}
    \vspace{2mm}
    \caption{Ablation study on three synthetic datasets.}
     \label{tab:ablation_study}
    \end{minipage}
    \hfill
    \begin{minipage}[t]{.45\textwidth}
    \centering
    \begin{tabular}{l|cc}
        \toprule
         & PSNR$\uparrow$ & SSIM$\uparrow$ \\ \midrule
        E2VID+SAM & 19.394 & 0.854\\
        GT binary mask & \textbf{21.189} & \textbf{0.878}\\
        Ours (Snake) & 20.656 & 0.869\\ 
        \bottomrule
    \end{tabular}
    \vspace{2mm}
     \caption{The impact of the binary mask quality on the final results.}
     \label{tab:binary_mask_quality}
    \end{minipage}
\end{table}

\subsection{Compared Methods}
\label{ssec:comparison_methods}
We compare with 3DGS~\cite{kerbl3Dgaussians} that performs static novel view synthesis, and Deformable-3DGS (D3DGS)~\cite{yang2024deformable}, an extension of 3DGS to dynamic scenes.
Since both methods require intensity frames and cannot operate on events, we reconstruct greyscale frames from event streams with E2VID \cite{rebecq2019high}.
The background colour of frames reconstructed with E2VID often differs from that of GT images and significantly impacts quantitative evaluation results.
For fair evaluation, we replace the background with a specific colour using masks generated by Segment Anything Model (SAM) \cite{kirillov2023segany}.

\subsection{Results and Comparisons}
\label{ssec:comparison}

In \cref{tab:quantitative_results_on_synthetic_dataset_full} and \cref{fig:qualitative_result_synthetic}, we show the results on the synthetic dataset.
\ours~performs comparably or better than other methods, both quantitatively and qualitatively.
Since 3DGS assumes a static scene, it fails to restore dynamic parts such as the arms in
Dynamic Lego and the human arms and legs in Jumping Jacks.
On the other hand, D3DGS performs quantitatively on par with the proposed method. However, in Jumping Jacks, details on the human face and clothing texture appear severely blurred. This is because the pixel values in frames reconstructed by E2VID are not always accurate. In particular, for rapidly moving objects, a large amount of event data spreads across a wide area in the frame, requiring the reconstruction of all pixel values in those regions, which increases the error.
Since our method directly processes the event stream, there is no information loss. 
This enables view synthesis that reflects the rich texture information of the object.

\cref{fig:qualitative_result_real} shows the qualitative results on the real dataset.
Our method accurately represents the appearance and shape of the deformed object captured by real event cameras.
This shows that our method works in real-world scenarios as well.
The other two methods fail to render due to the poor performance of frame reconstruction with E2VID.

\subsection{Ablations}
\label{ssec:ablation}

\noindent \textbf{Silhouette Loss and Fine-tuning in Fine Stage.}
We conduct an ablation study to validate the effect of the silhouette loss and fine-tuning the coarse deformation model in Fine Stage.
\cref{tab:ablation_study} shows the ablation study results.
We find that fine-tuning the coarse model in the second stage significantly improves the performance.
In the coarse stage, the model is primarily supervised with binary masks, which may contain unexpected point motion, and fine-tuning effectively suppresses it.
In addition, the silhouette loss $\mathcal{L}_\mathrm{sil}$ in the fine stage is effective in improving image quality by suppressing unwanted 3D Gaussians in the background area.

\noindent \textbf{Binary Mask Quality.}
Additionally, we evaluate the impact of binary mask quality on the final results. Specifically, we compare the performance of the proposed method when using binary masks generated with Snake, binary masks generated with SAM on frames reconstructed from events (E2VID+SAM), and ground-truth (GT) binary masks. The results are presented in \cref{tab:binary_mask_quality}. The performance of the proposed method is shown to be sensitive to the quality of the binary masks. As noted earlier, image restoration using E2VID often fails, resulting in lower-quality binary masks derived with SAM
from the reconstructed images. In contrast, binary masks generated with Snake, which does not require intermediate frame reconstruction,  contribute more effectively to the final results than those obtained with E2VID+SAM, although their performance remains inferior to that achieved with GT binary masks. 
These findings suggest that the proposed method could achieve further improvements if more accurate binary masks can be generated directly from events.

\begin{figure}
    \centering
    \includegraphics[width=1.0\linewidth]{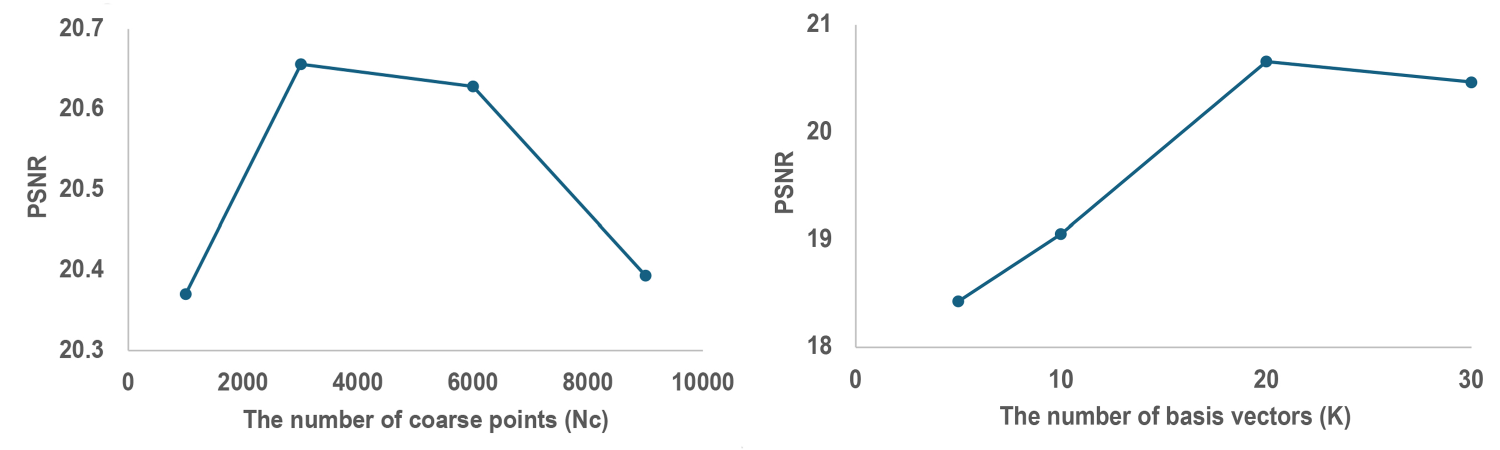}
    \caption{Hyperparameter selection and the average performance of the proposed method on three synthetic sequences.} 
    \label{fig:hyperparameter_selection}
\end{figure}

\noindent \textbf{Hyperparameter Selection.}
Finally, we discuss the selection of hyperparameters. \cref{fig:hyperparameter_selection} illustrates how performance varies with changes in the number of coarse points $N_c$ and the number of basis vectors $K$. In particular, variations in $K$ have a substantial impact on the final performance. While increasing $K$ enhances the expressive capacity for motion, it also appears to increase the difficulty of optimization.

\section{Conclusions}
\label{sec:conclusion}

This paper, for the first time, successfully studies and reconstructs novel views of a non-rigid object with large deformations at high temporal resolutions observed by a single monocular event camera. 
The demonstrated results show high effectiveness of the proposed 3DGS-based model and high accuracy on the level expected in our challenging setting. 
Ev4DGS outperforms competing methods applicable in our setting by $10\%$ in PSNR, which highlights the advantage of using events directly without intermediate frame reconstructions. 
Moreover, our method can work with brightness and colour events. 
 
We believe event cameras have a high potential in 3D reconstruction of non-rigid objects due to their characteristics. 
Despite achieving the highest accuracy compared to the tested baselines, one limitation of our model is the appearance artefacts noticeable as individual or groups of rendered Gaussians. 
The proposed method is the first step towards accurate novel-view rendering of non-rigid objects from event streams, opening up multiple avenues for future research. 
We hope the proposed method will be one promising example showing the effectiveness of event cameras in novel-view rendering of non-rigid objects.

\paragraph{Acknowledgements.} 
This work was partially supported by 
Japan Society for the Promotion of Science, KAKENHI, Grant Number JP23H03422; JSPS Overseas Challenge Program for Young Researchers; JST BOOST, Grant Number JPMJBS2409.

\bibliography{egbib}

\newpage
\appendix
\setcounter{figure}{0}
\setcounter{table}{0}
\setcounter{section}{0}
\renewcommand{\thesection}{\Alph{section}}
\renewcommand*\thetable{\Roman{table}}
\renewcommand*\thefigure{\Roman{figure}}

\bmvaResetAuthors

\title{Ev4DGS: Novel-view Rendering of Non-Rigid Objects from Monocular Event Streams (Supplementary Material)}

\addauthor{Takuya Nakabayashi}{nakka0204@keio.jp}{1}
\addauthor{Navami Kairanda}{nkairand@mpi-inf.mpg.de}{2}
\addauthor{Hideo Saito}{hs@keio.jp}{1}
\addauthor{Vladislav Golyanik}{golyanik@mpi-inf.mpg.de}{2}

\addinstitution{
 Keio University\\
 Yokohama, Japan
}
\addinstitution{
 Max Planck Institute for Informatics\\
 Saarbrücken, Germany
}

\maketitle

\section{Implementation Details}
\label{sec:implementation_supp}
In the coarse stage, we train the model for $200k$ iterations.
We use the cosine annealing scheduler to update the learning rates for the basis weights $\alpha_k(t)$ and basis vectors $\{\mathbf{B}_k\}_{k=1}^K$.
We set K to $20$.
In the fine stage, we optimise parameters in 3D Gaussians for $40k$ iterations.
We also finetune the coarse model after 3k iterations in this stage.
The loss weights are $\lambda_e=1, \lambda_s=1$.
The number of coarse points is set to $N_c=3000$, and the number of Gaussians at convergence, $N_g \approx 50k$. 
Note that $N_c \ll N_g$, therefore, the coarse model can be trained for longer without much performance drop. 
We perform our experiments on NVIDIA Quadro GV100, and the full reconstruction takes ${\approx}4$ hours. 

\section{Metrics}
Since events only provide information about changes in luminance, it is not possible to estimate absolute luminance values from event data. 
As a result, the images rendered by both the proposed and baseline methods may exhibit luminance shifts relative to the ground truth, potentially leading to unfair evaluations. 
To address this issue, colour correction is applied to all images prior to computing evaluation metrics, following the approach adopted in previous works~\cite{rudnev2023eventnerf}.
To align the rendered images $\{I_n\}_{n=1}^N$ with the ground truth images $\{G_n\}_{n=1}^N$, we estimate the optimal scale $\hat{s}$ and bias $\hat{b}$ as follows:
\begin{equation}
    \begin{split}
        f(I)&=\exp(s\log(I)+b),\\
        \hat{s},\hat{b}&=\argmax_{s,b} \frac{1}{N}\sum_{n=1}^N(PSNR(f(I_n),G_n)),
    \end{split}
\end{equation}
where $PSNR(\cdot,\cdot)$ is a function to compute PSNR value for two images.
We use Adam optimizer~\cite{kingma2014adam} in the experiments.
Then, we compute PSNR and SSIM values for the transformed images ${f(I_n)}_{n=1}^{N}$.

\section{Datasets}
\label{sec:datasets_supp}

\noindent
\textbf{Synthetic Dataset.}
To create the synthetic dataset, we first render RGB frames in 4k fps with Blender.
At the same time, the camera parameters at each timestamp are also recorded.
The camera moves around the object, always facing towards the object.
The rendered animation is then provided as input to ESIM \cite{rebecq2018esim} for event generation.
Though ESIM can only simulate greyscale events, our synthetic dataset contains RGB events.
We generate events for each colour channel independently and integrate them as post-processing to simulate RGB events.

\begin{figure*}[h!]
    \centering
    \includegraphics[width=0.4\linewidth]{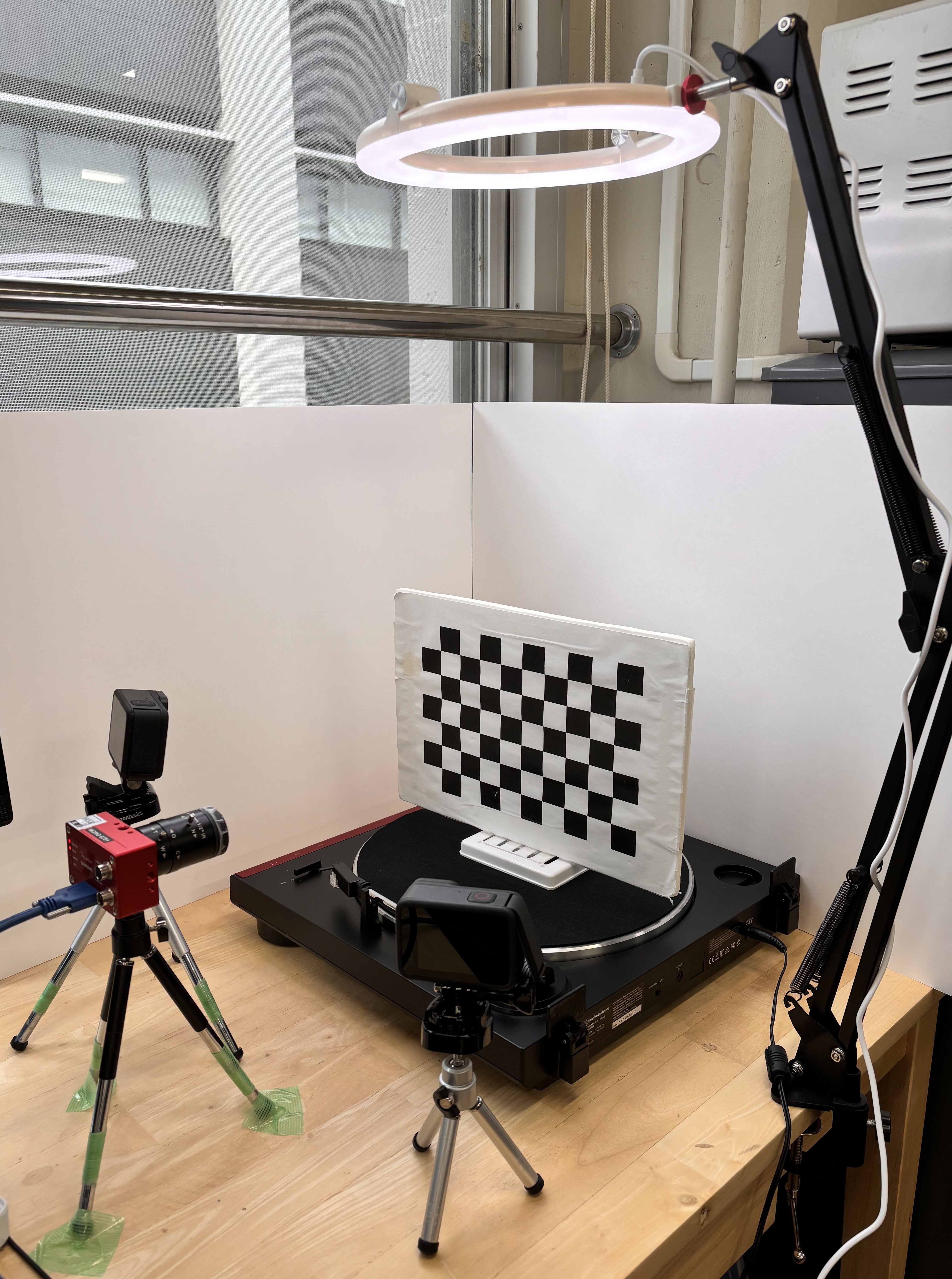}
    \caption{Real data recording setup.} 
    \label{fig:real_dataset_recoarding}
\end{figure*}

\noindent
\textbf{Real Dataset.}
The recording process of the real dataset is the same as EventNeRF \cite{rudnev2023eventnerf}.
\cref{fig:real_dataset_recoarding} shows the setup for the recording of the real dataset.
The event camera used is a DAVIS 346 MONO, and the RGB camera is a GoPro HERO10 Black. 
The data captured by the RGB camera is used solely for evaluation purposes and not for training.
Calibration between the event camera and the RGB camera is performed using a checkerboard pattern placed on a turntable. 
The relative positions and poses of the event camera and RGB camera are then estimated.
Subsequently, the subject is placed on the turntable, which is rotated at a constant speed. 
For temporal synchronization between the event camera and the RGB camera, a light is flashed in front of the cameras, and the captured timings are aligned accordingly.

\end{document}